\documentclass[10pt,twocolumn,letterpaper]{article}

\makeatletter
\let\paperdir\@empty
\ifx\CurrentFilePath\@empty
  \IfFileExists{\jobname.fls}{%
    \RequirePackage{currfile-abspath}%
    \getmainfile
    \expandafter\filename@parse\expandafter{\themainfile}%
    \edef\paperdir{\filename@area}%
  }{}
\else
  \edef\paperdir{\CurrentFilePath/}
\fi
\ifx\paperdir\@empty
  \IfFileExists{./main_arXiv.tex}{\def\paperentry{main_arXiv.tex}}{\let\paperentry\@empty}
  \ifx\paperentry\@empty
    \RequirePackage{graphicx,amsmath}
    \ExplSyntaxOn
    \ior_open:Nn \g_tmpa_ior {\jobname.log}
    \ior_str_map_inline:Nn \g_tmpa_ior {
      \regex_extract_once:nnNT { \A\*\*\s*"?([^"]+\.tex)"?\s*\Z } {#1} \l_tmpa_seq {
        \edef\paperentry{\seq_item:Nn \l_tmpa_seq {2}}
        \ior_map_break:
      }
    }
    \ior_close:N \g_tmpa_ior
    \ExplSyntaxOff
    \ifx\paperentry\@empty
      \PackageError{Manuscript}{Cannot locate the manuscript directory}
        {Compile with latexmk or from the directory containing main_arXiv.tex.}
    \fi
    \expandafter\filename@parse\expandafter{\paperentry}
    \let\paperdir\filename@area
  \fi
\fi
\ifx\paperdir\@empty\else
  \declare@file@substitution{cvpr.sty}{\paperdir cvpr.sty}
  \declare@file@substitution{supplment.tex}{\paperdir supplment.tex}
  \declare@file@substitution{theory_geometry.tex}{\paperdir figures/theory_geometry.tex}
\fi
\makeatother
\usepackage[pagenumbers]{cvpr}
\usepackage{amsmath,amssymb,amsthm,bm,mathtools}
\usepackage{graphicx,booktabs,multirow,array}
\usepackage{microtype}
\usepackage{tikz}
\usetikzlibrary{arrows.meta,positioning,calc}
\definecolor{cvprblue}{rgb}{0.21,0.49,0.74}
\usepackage[breaklinks,colorlinks,allcolors=cvprblue]{hyperref}
\usepackage[capitalize,noabbrev]{cleveref}
\definecolor{navy}{RGB}{32,66,104}
\definecolor{teal}{RGB}{0,112,112}
\definecolor{amber}{RGB}{165,98,16}
\newtheorem{theorem}{Theorem}

\makeatletter
\ifx\paperdir\@empty\else
  \let\paper@readgraphics\Ginclude@graphics
  \def\Ginclude@graphics#1{\paper@readgraphics{\paperdir#1}}
  \let\paperbibdir\paperdir
  \ExplSyntaxOn
  \tl_if_in:NnT \paperdir { ~ } {
    \RequirePackage[enquotefilenames]{verbatimcopy}
    \OldVerbatimCopy{\paperdir references.bib}{\jobname-paper-references.bib}
    \OldVerbatimCopy{\paperdir ieeenat_fullname.bst}{\jobname-paper-ieeenat_fullname.bst}
    \def\paperbibdir{\jobname-paper-}
  }
  \ExplSyntaxOff
  \let\readpaperbib\bibliography
  \let\readpaperbst\bibliographystyle
  \renewcommand*{\bibliography}[1]{\readpaperbib{\paperbibdir#1}}
  \renewcommand*{\bibliographystyle}[1]{\readpaperbst{\paperbibdir#1}}
\fi
\providecommand*{\@LN@col}[1]{}
\providecommand*{\@LN}[2]{}
\renewcommand\paragraph{\@startsection{paragraph}{4}{\z@}%
  {2.0ex \@plus .5ex \@minus .2ex}%
  {-1em}%
  {\normalfont\normalsize\bfseries}}
\makeatother
\newcommand*{\papertitle}{Backbone-Specific LoRA Geometry:\texorpdfstring{\\}{ }%
Orthogonality, Superposition, and Provenance}
\title{\papertitle}
\newcommand*{\affmark}[1]{\texorpdfstring{\textsuperscript{#1}}{}}
\newcommand*{\paperauthors}{Kazuhiko Kawamoto\affmark{1}, Keigo Watanabe\affmark{1}, and Hiroshi Kera\affmark{1,2}}
\author{\paperauthors\\[3pt]
  {\normalsize \affmark{1}Chiba University\qquad \affmark{2}National Institute of Informatics}}
\hypersetup{pdftitle={\papertitle},pdfauthor={\paperauthors}}
\begin{document}
\maketitle
\begin{abstract}
Sharing one low-rank adaptation (LoRA) update across pretrained models
simplifies deployment.
Composition methods typically combine updates from a common backbone,
while cross-model transfer transforms updates for a target backbone.
The conditions for sharing updates across backbones while preserving
accuracy and distinguishing individual contributions remain unclear.
Our idea is to use backbone-specific update directions as signatures of
their training backbones.
We derive a cross-backbone correlation bound from attention symmetries
and equivariant LoRA learning under independent, uniform pretrained
feature orientations.
A local gradient-flow model explains positive mean correlation between
independent runs on the same backbone and task.
These patterns motivate cosine-based backbone identification and
inner-product tests for registered-adapter reuse without coordinate alignment.
Experiments on CLIP backbones show that shared sums achieve nearly the
same mean accuracy as separate adapters.
Cosine scores reliably identify source backbones across the evaluated tasks.
Registered adapters remain detectable in sums of adapters from different backbones.
\end{abstract}
\section{Introduction}
Supporting the same image-classification task with several pretrained models
creates a practical maintenance problem.
Each model needs a compatible task-specific update, and these updates must
be stored and distributed with information about their training backbones.
A single shared update would simplify maintenance and distribution,
provided the shared update preserves each model's adapted accuracy.
When adapters are exchanged or combined, users may also need to verify their
training sources from the weights because metadata are missing.

A common solution is to adapt each backbone separately with low-rank adaptation
(LoRA)~\cite{hu2022lora} and deploy the matching update.
Training low-rank updates reduces the amount of task-specific storage
but retains separate adapters and their associations with backbones.
To reuse existing adaptations, composition methods combine updates primarily
around a shared pretrained model~\cite{ilharco2023task,yadav2023ties},
while cross-model transfer methods transform an update for a recipient
backbone~\cite{farhadzadeh2025lorax,kawamoto2025orthogonal}.
Combining updates also raises the problem of interference, addressed through
constrained subspaces~\cite{zhang2025osrm} or attention symmetries for model
fusion~\cite{zhang2025rotation}.
Other methods identify training sources by examining model responses,
checkpoint weights, or deliberately embedded
identifiers~\cite{he2025routemark,liang2025origintracer,xu2025lorafp}.
An open question remains: under what conditions can adapters trained on
different backbones be shared while their individual contributions
remain distinguishable?

The key idea is to retain backbone-specific update directions in a shared sum
and use these directions to trace contributions.
Prior superposition work sums nearly orthogonal same-task updates across
backbones in their original coordinates~\cite{lora_superposition}.
The sum can serve several backbones if each backbone responds mainly to its
own component and foreign components cause little prediction change.
Near-orthogonal directions reduce cross terms, so inner products with
registered adapters estimate their contributions to the sum.
For backbone identification, the method compares a query with reference
adapters and selects the backbone with the highest average cosine similarity.
To explain this geometry, we derive a cross-backbone correlation bound from
attention symmetries and equivariant LoRA learning under independent,
uniform pretrained feature orientations.
A local gradient-flow analysis explains positive mean correlation between
independent runs on one backbone for the same task.

On seven ViT-B/16 CLIP backbones and eight datasets, prior results show that
superposition largely preserves individual-adapter
accuracy~\cite{lora_superposition}.
We measure correlations within and across these backbones and test
identification and detection across training runs.
Independently trained same-backbone, same-task adapters serve as hard negatives
to distinguish adapter reuse from shared training origins.

Our contributions are:
\begin{itemize}
\item A conditional theory of cross-backbone LoRA orthogonality,
with a second-moment bound from attention symmetries
and equivariant LoRA learning, and a local analysis of
within-backbone correlation.
\item Correlation-based backbone identification and
registered-adapter detection, evaluated across independent training runs.
\end{itemize}
\section{Related Work}
\paragraph{Adapter composition and transfer.}
LoRA~\cite{hu2022lora} represents adaptation as low-rank increments to
frozen pretrained weights.
Task arithmetic~\cite{ilharco2023task}, TIES~\cite{yadav2023ties},
and DARE~\cite{yu2024dare} combine task updates,
while LoRAHub~\cite{huang2024lorahub} and Arrow~\cite{ostapenko2024arrow}
reuse collections of adapters.
These approaches principally study adaptation and composition around a shared
pretrained backbone.
Cross-model transfer addresses a different obstacle: updates learned on
separately pretrained models may use incompatible internal coordinates.
Subspace projection~\cite{farhadzadeh2025lorax},
re-basing~\cite{rinaldi2025transfusion}, and few-shot orthogonal
alignment~\cite{kawamoto2025orthogonal} are approaches to transporting such updates.
LoRA superposition~\cite{lora_superposition} retains the original updates
and combines adapters already trained on each participating backbone.
We analyze cross-backbone update correlations and extend their use to backbone
identification and adapter detection.

\paragraph{Orthogonality and symmetry.}
The rank-and-seed analysis in the original LoRA paper~\cite{hu2022lora}
examines singular subspaces of the ${\bm{A}}$ factor on one pretrained model;
we compare effective ${\bm{B}}{\bm{A}}$ updates across backbones
using cosine similarity.
O-LoRA~\cite{wang2023olora} maintains orthogonal task subspaces for continual
learning, and OSRM~\cite{zhang2025osrm} constrains LoRA subspaces before
training to reduce data-dependent merging interference.
We instead study near-orthogonality that appears without an orthogonality objective.
Parameter superposition~\cite{cheung2019superposition} and task singular
vectors~\cite{gargiulo2025tsv} also use parameter-space directions
for model combination.
Attention rotation symmetry~\cite{zhang2025rotation} supplies function-preserving
changes of internal coordinates.
We use signed-permutation symmetry to derive a conditional bound on
update correlations.
The application methods keep adapters in their original coordinates,
and functional noninterference is evaluated separately.

\paragraph{Adapter provenance.}
Adapter provenance methods identify training sources or reused adapters from
weights, outputs, or deliberately embedded identifiers.
Origin Tracer~\cite{liang2025origintracer} compares candidate and base models
to infer whether one was obtained by LoRA fine-tuning the other;
centered residual signatures~\cite{thakur2026traces} likewise
analyze model checkpoints.
Behavioral fingerprints use refusal directions~\cite{xu2026refusal} or expert
routing responses~\cite{he2025routemark}, the latter enabling
attribution within MoE merges.
LoRA-FP~\cite{xu2025lorafp} embeds fingerprints during adaptation,
while LoRA-Key~\cite{wang2026lorakey} trains a reusable watermark adapter
for composition with target adapters.
Our detectors inspect unmodified adapters.
W2T~\cite{han2026w2t} learns adapter representations after removing the
ambiguity of low-rank factorization through QR-SVD canonicalization.
Inner products between effective updates remove factorization ambiguity
without changing backbone coordinates.
We distinguish identifying a training backbone from detecting a specific
registered adapter in a sum; independently retrained adapters from the same
backbone and task provide hard negatives.
\section{Geometry of LoRA Updates}
\label{sec:theory}
\begin{figure*}[t]
\centering
\begin{tikzpicture}[x=1cm,y=.94cm,font=\footnotesize,
  flow/.style={-{Latex[length=1.6mm]},line width=.65pt,draw=navy},
  stage/.style={draw=navy!35,fill=navy!4,rounded corners=1.5pt,
    align=center,text width=1.9cm,minimum height=.72cm,inner sep=3pt}]
\useasboundingbox (0,.75) rectangle (17.05,6.28);
\draw[gray!35] (8.5,.80)--(8.5,6.2);

\node[anchor=north west,font=\small\bfseries] at (.15,6.18)
  {(a) Learning follows coordinates};
\node[align=center,text width=7.7cm] at (4.2,5.53)
  {One backbone, two equivalent representations};
\node[stage] (a1) at (1.6,4.48) {Frozen\\backbone};
\node[stage] (a2) at (6.65,4.48) {Reference update\\$\bm\tau_i'$};
\node[stage] (a3) at (1.6,2.43) {Equivalent\\backbone};
\node[stage] (a4) at (6.65,2.43) {Transformed update\\$\bm\tau_i$};
\draw[flow] (a1)--node[above] {LoRA training} (a2);
\draw[flow] (a3)--node[above] {LoRA training} (a4);
\draw[flow] (a1)--node[left,align=center,font=\scriptsize]
  {permute\\and flip\\features} (a3);
\draw[flow] (a2)--node[right] {$\bm P_i$} (a4);
\node[align=center,text width=3.55cm,font=\scriptsize] at (4.2,3.48)
  {Corresponding initialization\\and training choices\\[2pt]
   $\bm{B}_0=\bm{0}$, same $\bm{A}_0$};
\node[align=center,text width=7.7cm] at (4.2,1.33)
  {$\bm\tau_i=\bm P_i\bm\tau_i'$\\[3pt]
   Either path gives the same learned update};

\node[anchor=north west,font=\small\bfseries] at (8.75,6.18)
  {(b) Correlations across backbones};
\node[align=center,text width=7.8cm] at (12.8,5.62)
  {Independent, uniform pretrained orientations};
\node[align=center,text width=7.8cm] at (12.8,5.20)
  {$\bm\tau_i=\bm P_i\bm\tau_i'$,\quad
   $\bm\tau_j=\bm P_j\bm\tau_j'$};
\node[font=\scriptsize] at (12.8,4.82)
  {Exact example probabilities (one block, one column)};
\definecolor{correlationBlue}{HTML}{3B73B9}
\definecolor{correlationRed}{HTML}{D65A5A}
\tikzset{
  correlation20/.style={fill=correlationBlue,fill opacity=.5,
    draw=correlationBlue!85!black,draw opacity=.85,line width=.2pt},
  correlation100/.style={fill=correlationRed,fill opacity=.5,
    draw=correlationRed!85!black,draw opacity=.85,line width=.2pt}}
\begin{scope}[xshift=.565cm,xscale=1.4]
\filldraw[correlation20] (6.58524,2.30000) rectangle (6.77476,2.30000);
\filldraw[correlation20] (6.79124,2.30000) rectangle (6.98076,2.30009);
\filldraw[correlation20] (6.99724,2.30000) rectangle (7.18676,2.30083);
\filldraw[correlation20] (7.20324,2.30000) rectangle (7.39276,2.30495);
\filldraw[correlation20] (7.40924,2.30000) rectangle (7.59876,2.32105);
\filldraw[correlation20] (7.61524,2.30000) rectangle (7.80476,2.36736);
\filldraw[correlation20] (7.82124,2.30000) rectangle (8.01076,2.46839);
\filldraw[correlation20] (8.02724,2.30000) rectangle (8.21676,2.63679);
\filldraw[correlation20] (8.23324,2.30000) rectangle (8.42276,2.84728);
\filldraw[correlation20] (8.43924,2.30000) rectangle (8.62876,3.02970);
\filldraw[correlation20] (8.64524,2.30000) rectangle (8.83476,3.10268);
\filldraw[correlation20] (8.85124,2.30000) rectangle (9.04076,3.02970);
\filldraw[correlation20] (9.05724,2.30000) rectangle (9.24676,2.84728);
\filldraw[correlation20] (9.26324,2.30000) rectangle (9.45276,2.63679);
\filldraw[correlation20] (9.46924,2.30000) rectangle (9.65876,2.46839);
\filldraw[correlation20] (9.67524,2.30000) rectangle (9.86476,2.36736);
\filldraw[correlation20] (9.88124,2.30000) rectangle (10.07076,2.32105);
\filldraw[correlation20] (10.08724,2.30000) rectangle (10.27676,2.30495);
\filldraw[correlation20] (10.29324,2.30000) rectangle (10.48276,2.30083);
\filldraw[correlation20] (10.49924,2.30000) rectangle (10.68876,2.30009);
\filldraw[correlation20] (10.70524,2.30000) rectangle (10.89476,2.30000);
\filldraw[correlation100] (7.63378,2.30000) rectangle (7.78622,2.30001);
\filldraw[correlation100] (7.83978,2.30000) rectangle (7.99222,2.30092);
\filldraw[correlation100] (8.04578,2.30000) rectangle (8.19822,2.32648);
\filldraw[correlation100] (8.25178,2.30000) rectangle (8.40422,2.57602);
\filldraw[correlation100] (8.45778,2.30000) rectangle (8.61022,3.40265);
\filldraw[correlation100] (8.66378,2.30000) rectangle (8.81622,4.04341);
\filldraw[correlation100] (8.86978,2.30000) rectangle (9.02222,3.40265);
\filldraw[correlation100] (9.07578,2.30000) rectangle (9.22822,2.57602);
\filldraw[correlation100] (9.28178,2.30000) rectangle (9.43422,2.32648);
\filldraw[correlation100] (9.48778,2.30000) rectangle (9.64022,2.30092);
\filldraw[correlation100] (9.69378,2.30000) rectangle (9.84622,2.30001);

\draw[->,gray!80,line width=.45pt] (6.55,2.3)--(10.98,2.3);
\draw[->,gray!80,line width=.45pt] (6.58,2.3)--(6.58,4.5);
\foreach \xx/\lab in {6.68/-1,7.71/-0.5,8.74/0,9.77/0.5,10.8/1}{
 \draw[gray!80] (\xx,2.3)--(\xx,2.24);
 \node[below,font=\scriptsize,inner sep=2pt] at (\xx,2.24) {$\lab$};}
\foreach \yy/\lab in {2.3/0,3.21111/0.2,4.12222/0.4}{
 \draw[gray!80] (6.58,\yy)--(6.52,\yy);
 \node[left,font=\scriptsize,inner sep=2pt] at (6.52,\yy) {$\lab$};}
\node[rotate=90,font=\scriptsize] at (6.02,3.4) {Probability per cosine bin};
\node[font=\scriptsize] at (8.75,1.81) {Cosine $C_{ij}$; $0$ means orthogonal};
\filldraw[correlation20] (9.46,4.32) rectangle (9.67,4.48);
\node[anchor=west,font=\scriptsize] at (9.73,4.40) {$d=20$};
\filldraw[correlation100] (9.46,4.03) rectangle (9.67,4.19);
\node[anchor=west,font=\scriptsize] at (9.73,4.11) {$d=100$};
\end{scope}
\node[align=center,text width=7.8cm] at (12.8,1.24)
  {Larger $d$: narrower distribution\\[2pt]
   $\mathbb{E}[C_{ij}^{2}]=1/d$ in this example};
\end{tikzpicture}%
\caption{Coordinate geometry in Secs.~\ref{sec:equivariance} and~\ref{sec:crossgeometry}.
The matrix $\bm P_i$ applies headwise signed permutations to the adapter vector.
(a) Equivariant LoRA training commutes with coordinate changes.
(b) For a single width-$d$ block with one column and
$\bm\tau_i'=\bm\tau_j'=(1,\ldots,1)^\top/\sqrt{d}$,
independent, uniform signed permutations yield cosine distributions that
concentrate around zero as $d$ increases, with $\mathbb{E}[C_{ij}^2]=1/d$.
}
\label{fig:theory_geometry}
\end{figure*}
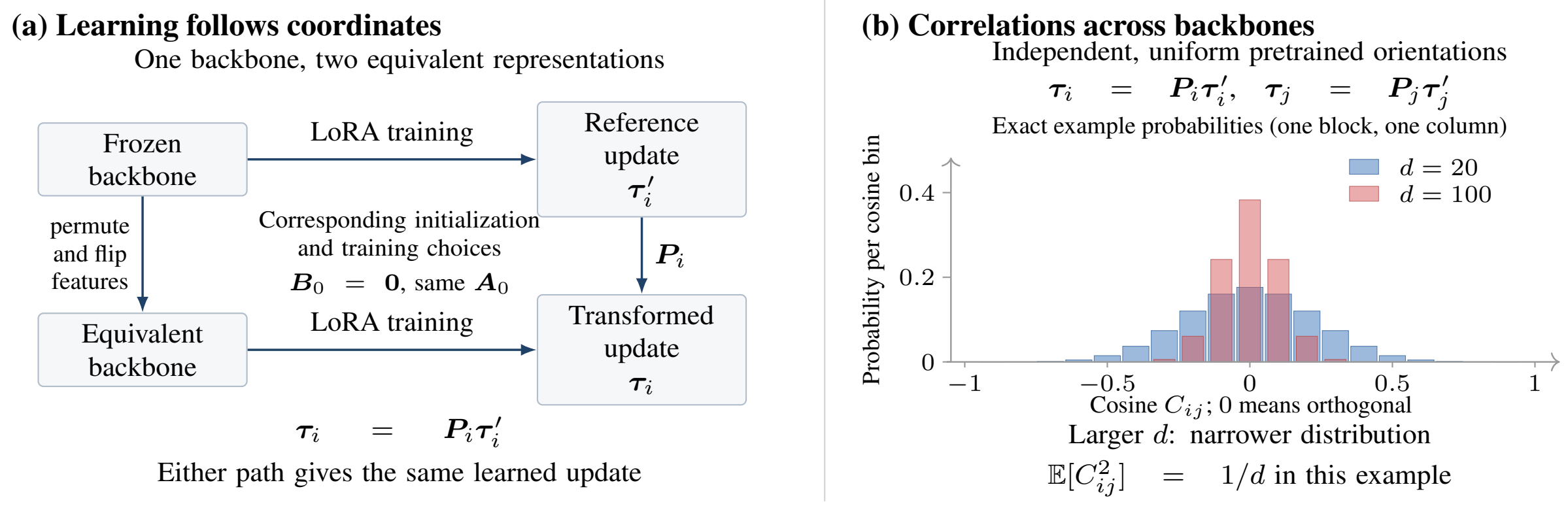

We derive sufficient conditions for same-task LoRA updates to be nearly
orthogonal across backbones.
Learned updates follow attention coordinate changes; independent, uniform
orientations then suppress cross-backbone correlations (\cref{fig:theory_geometry}).
A local gradient-flow model explains positive mean correlation within one backbone.

\paragraph{Notation.}
Bold symbols denote vectors and matrices.
Vectors use Euclidean inner products and norms.
For equal-sized matrices, the Frobenius inner product and norm are
\begin{equation}
\left\langle \bm{M},\bm{N}\right\rangle_F=\operatorname{tr}(\bm{M}^{\top}\bm{N}),
\quad \left\lVert \bm{M}\right\rVert_F=\sqrt{\left\langle \bm{M},\bm{M}\right\rangle_F}.
\label{eq:frobenius}
\end{equation}
Here $\operatorname{tr}$ denotes the trace.

\subsection{Attention symmetries and LoRA learning}
\label{sec:equivariance}
Fix a task and one adaptation run per backbone, including initialization
and stochastic choices.
For backbone $i$, each Q/V module $\ell=1,\ldots,J$ has effective update
\begin{equation}
\Delta {\bm{W}}_{i,\ell}=\frac{\alpha}{r}{\bm{B}}_{i,\ell}{\bm{A}}_{i,\ell}\in\mathbb{R}^{n_\ell\times m_\ell}.
\label{eq:lora}
\end{equation}
Here $\bm{B}\in\mathbb{R}^{n_\ell\times r}$
and $\bm{A}\in\mathbb{R}^{r\times m_\ell}$ are the trainable factors,
$r$ is their rank parameter, and $\alpha/r$ is the update scale.
We compare effective updates because different factors give the same product:
$\bm B\bm A=(\bm B\bm C)(\bm C^{-1}\bm A)$ for any invertible $\bm C$.

Concatenate the vectorized Q/V updates into one adapter vector:
\begin{equation}
\bm\tau_i=
\begin{bmatrix}
\operatorname{vec}(\Delta\bm{W}_{i,1})\\
\vdots\\
\operatorname{vec}(\Delta\bm{W}_{i,J})
\end{bmatrix}.
\label{eq:adaptervector}
\end{equation}
Here $\operatorname{vec}$ stacks matrix columns; concatenation uses the
same module order for all adapters.
Vectorization preserves inner products:
$\bm\tau_i^\top\bm\tau_j=
\sum_{\ell=1}^J\langle\Delta\bm{W}_{i,\ell},\Delta\bm{W}_{j,\ell}\rangle_F$.
For a nonzero adapter, write
$\hat{\bm\tau}_i=\bm\tau_i/\lVert\bm\tau_i\rVert_2$;
normalization uses the whole adapter, not individual modules.

For an ordinary dot-product attention head of width $d_h$ with column
activations, the following transformations preserve its output for
orthogonal $\bm R,\bm T\in\mathbb R^{d_h\times d_h}$~\cite{zhang2025rotation}:
\begin{equation}
\begin{split}
{\bm{W}}_Q'&={\bm{R}}{\bm{W}}_Q,\quad {\bm{W}}_K'={\bm{R}}{\bm{W}}_K,\\
{\bm{W}}_V'&={\bm{T}}{\bm{W}}_V,\quad {\bm{W}}_O'={\bm{W}}_O{\bm{T}}^\top
\end{split}
\label{eq:gauge}
\end{equation}
For query and key vectors ${\bm{q}},{\bm{k}}$,
$({\bm{R}}{\bm{q}})^\top({\bm{R}}{\bm{k}})={\bm{q}}^\top {\bm{k}}$,
so the attention weights are unchanged.
For the aggregated value $\bm{h}$, the output projection cancels
the value-side change:
$(\bm{W}_O\bm{T}^\top)(\bm{T}\bm{h})=\bm{W}_O\bm{h}$.
Here ${\bm{W}}_O$ contains this head's output-projection columns;
biases transform consistently.
Applying these transformations independently in each head preserves the
backbone function. Adaptation trains only Q/V LoRA factors and freezes
all pretrained weights.

Learning is \emph{equivariant} if a coordinate change before training gives
the same update as training first and then applying that change
(\cref{fig:theory_geometry}(a)).
Since AdamW updates each coordinate separately, we restrict $\bm R,\bm T$
to \emph{signed permutations}: coordinate reorderings and sign flips.
Both representations use $\bm B_0=\bm0$, the same $\bm A_0$, and zero
optimizer states.
Equivariance holds if hyperparameters match across permuted coordinates,
mini-batches match, and stochastic and additional head operations respect
the transformation (Supplementary Sec.~\ref{app:equivariance}).
The whole-adapter update therefore satisfies
\begin{equation}
\bm\tau_i=\bm P_i\bm\tau_i'.
\label{eq:orientation}
\end{equation}
Here $\bm\tau_i'$ is the learned update in a reference coordinate
representation of backbone $i$, and $\bm P_i$ is the signed permutation
induced on the full vector by the headwise transformations.

\subsection{Bounding correlations across backbones}
\label{sec:crossgeometry}
We bound correlations between $\bm\tau_i$ and $\bm\tau_j$ by separating
entries according to attention head.
Each attention head forms its own Q/K/V vectors from the full input features.
For example, a $768\times768$ Q/V update with 12 heads has twelve
$64\times768$ row blocks, all belonging to the same learned matrix.
Let $b$ index these head row blocks across all Q/V modules,
and let $\bm\tau_{i,b}$ contain the entries of block $b$, stacked column by column.
These subvectors partition $\bm\tau_i$, giving
$\lVert\bm\tau_i\rVert_2^2=\sum_b\lVert\bm\tau_{i,b}\rVert_2^2$.
For head width $d_b$, the row transformation
$\bm P_{i,b}\in\mathbb R^{d_b\times d_b}$ uses the same permutation
and signs in every column. These headwise actions induce $\bm P_i$
in \cref{eq:orientation}.

\paragraph{Assumptions.}
\emph{(A1) Equivariant adaptation.}
Under the conditions of Sec.~\ref{sec:equivariance}, changing head coordinates
before LoRA training gives the same update as applying that coordinate
change after training, as in \cref{eq:orientation}.
\emph{(A2) Independent, uniform orientations.}
Fix the reference pretrained models, task, and adaptation choices,
including initialization and stochastic training streams.
Conditional on these quantities, $\bm P_{i,b}$ are independent across
backbones and blocks, and uniform over all row permutations and sign patterns.
The reference updates $\bm\tau_i'$ are then fixed, and only the orientations
are random; no independence between reference updates is required.
Attention symmetry alone does not imply (A2);
Supplementary Sec.~\ref{app:orientation} gives sufficient pretraining conditions.

\begin{theorem}[Cross-backbone second moment]
\label{thm:orthogonality}
Fix two distinct backbones $i\ne j$ with nonzero updates, and
assume (A1)--(A2).
Let
$w_{i,b}=\lVert\bm\tau_{i,b}\rVert_2^2/\lVert\bm\tau_i\rVert_2^2$
be the fraction of squared update norm in block $b$.
These fractions are unchanged by the signed permutations.
The full-adapter cosine is
\begin{equation}
C_{ij}=\hat{\bm\tau}_i^\top\hat{\bm\tau}_j.
\label{eq:blockcosine}
\end{equation}
Then
\begin{align}
\mathbb E[C_{ij}]&=0,\\
\mathbb E[C_{ij}^2]&\leq\sum_b\frac{w_{i,b}w_{j,b}}{d_b}
\eqqcolon\kappa_{ij}.
\label{eq:energybound}
\end{align}
Consequently, for $\epsilon>0$,
\begin{equation}
\mathbb{P}(|C_{ij}|\geq\epsilon)\leq
\min\{1,\kappa_{ij}/\epsilon^2\}.
\label{eq:tail}
\end{equation}
Here $\mathbb{E}$ and $\mathbb{P}$ denote expectation and probability over
the signed permutations, conditional on $\bm\tau_i'$ and $\bm\tau_j'$.
\end{theorem}
\begin{proof}[Proof sketch]
Write $\bm\tau_i^\top\bm\tau_j=\sum_b Z_b$, where
$Z_b=\bm\tau_{i,b}^\top\bm\tau_{j,b}$.
Uniform signs give $\mathbb E[Z_b]=0$.
Averaging over row permutations and applying Cauchy--Schwarz gives
\begin{equation}
\mathbb E[Z_b^2]\leq
\frac{\lVert\bm\tau_{i,b}\rVert_2^2\lVert\bm\tau_{j,b}\rVert_2^2}{d_b}.
\label{eq:blockmomentbound}
\end{equation}
Independent block orientations make the variances add.
Dividing by $\lVert\bm\tau_i\rVert_2^2\lVert\bm\tau_j\rVert_2^2$
gives \cref{eq:energybound}; Markov's inequality gives \cref{eq:tail}.
Supplementary Sec.~\ref{app:moment} derives the exact second moment from the block matrices.
\end{proof}

\Cref{fig:theory_geometry}(b) illustrates the dependence on head width.
Equal squared norms across $L$ width-$d$ blocks give
$w_{i,b}=w_{j,b}=1/L$ and $\kappa_{ij}=1/(Ld)$.
With 12 layers, 12 heads, and Q/V projections, $L=288$ and $d=64$,
so $\sqrt{\mathbb{E}[C_{ij}^2]}\leq1/\sqrt{Ld}\approx0.0074$.
Unequal block energies can weaken this illustrative bound.

Under (A1)--(A2), the bound makes $|C_{ij}|\ge\epsilon$ unlikely
when $\kappa_{ij}/\epsilon^2$ is small.
Independent runs on one backbone share the same orientation,
so this argument does not bound their correlations.

\subsection{Correlations within a backbone}
\label{sec:withingeometry}
Independent runs on one backbone share an initial task gradient.
We use this common descent direction to explain positive mean cosine
in a local model.
Fix the backbone, task, and training data, and suppress the backbone index.
Let $\mathcal L(\bm\tau)$ be the full-batch loss and
$\bm g=\nabla_{\bm\tau}\mathcal L(\bm0)\ne\bm0$
its gradient at zero update.
Keep $\bm g$ fixed and consider the first-order loss term
$\Phi=\bm g^\top\bm\tau$.
The update vector $\bm\tau$ is formed from the LoRA products in \cref{eq:lora}.
We evolve the factors by unregularized gradient flow:
\begin{equation}
\dot{\bm A}_\ell=-\nabla_{\bm A_\ell}\Phi,
\quad
\dot{\bm B}_\ell=-\nabla_{\bm B_\ell}\Phi,
\label{eq:factorflow}
\end{equation}
where a dot denotes a time derivative.
Each run starts with $\bm B_\ell(0)=\bm0$ and draws the entries of
$\bm A_\ell(0)$ independently from distributions with densities.
The resulting adapter trajectory obeys
\begin{equation}
\begin{split}
\bm g^\top\dot{\bm\tau}=\dot\Phi
=-\sum_\ell\bigl(\lVert\nabla_{\bm A_\ell}\Phi\rVert_F^2
+\lVert\nabla_{\bm B_\ell}\Phi\rVert_F^2\bigr)
\leq0.
\end{split}
\label{eq:flowdescent}
\end{equation}
The nonzero gradient and this initialization make the initial derivative
strictly negative almost surely.
Since $\bm\tau_s(0)=\bm0$, the update from initialization $s$ satisfies
$\bm g^\top\bm\tau_s(t)<0$ almost surely for any fixed $t>0$.

Each run thus has a positive projection onto the same direction $-\bm g$.
Consequently, $\bm\mu(t)=\mathbb E_s[\hat{\bm\tau}_s(t)]\ne\bm0$.
For independent, identically distributed initializations $s,s'$,
\begin{equation}
\mathbb E_{s,s'}\!\left[
\hat{\bm\tau}_s(t)^\top\hat{\bm\tau}_{s'}(t)
\right]=\lVert\bm\mu(t)\rVert_2^2>0.
\label{eq:withinmean}
\end{equation}
Supplementary Sec.~\ref{app:withinflow} completes the proof.
The mean is over initialization on a fixed backbone; individual pairs need
not have positive cosine. The local model does not guarantee positive mean
cosine for final AdamW updates or an identification margin.
A shared coordinate change preserves within-backbone cosines,
consistent with Sec.~\ref{sec:crossgeometry}.
\section{Applications}
\label{sec:applications}
We study three applications: sharing one summed adapter across backbones,
identifying the backbone used to train an adapter, and detecting a registered
adapter in a combined update.
We use the whole-adapter vectors defined in \cref{eq:adaptervector}.

\subsection{Shared superposition}
\label{sec:sharing}
The goal is one shared adapter that preserves each backbone's adapted accuracy.
\emph{Solo} applies the target backbone's own adapter;
\emph{Exact sum} adds same-task adapters from all contributors.
For $K=|S|$ backbones with matching Q/V module shapes, form
\begin{equation}
\bm\tau_S=\sum_{i\in S}\bm\tau_i.
\label{eq:sum}
\end{equation}
The same effective update is applied unchanged to every $i\in S$.
At module $\ell$, with $c=\alpha/r$, the sum admits the factorization
\begin{equation}
\Delta\bm{W}_{S,\ell}
=c\,[\bm{B}_{1,\ell}\ \cdots\ \bm{B}_{K,\ell}]
\begin{bmatrix}\bm{A}_{1,\ell}\\ \vdots\\ \bm{A}_{K,\ell}\end{bmatrix},
\label{eq:concat}
\end{equation}
where the contributors are indexed $1,\ldots,K$.
The rank is at most $Kr$, and each term retains its original scale $c$.

\paragraph{Rank reduction by SVD.}
Exact superposition alone does not reduce storage: the factors in
\cref{eq:concat} require $Kr(n_\ell+m_\ell)$ entries per module, the same as
$K$ separate adapters.
For compression, we apply singular value decomposition (SVD) to each
effective update,
$\Delta\bm W_{S,\ell}=\bm U_\ell\bm\Sigma_\ell\bm V_\ell^\top$,
and retain the leading $r_c$ singular components:
\begin{equation}
\Delta\bm W_{S,\ell}^{(r_c)}
=\bm U_{\ell,r_c}\bm\Sigma_{\ell,r_c}\bm V_{\ell,r_c}^{\top}.
\label{eq:svdtruncation}
\end{equation}
The factors $\bm U_{\ell,r_c}\bm\Sigma_{\ell,r_c}$ and
$\bm V_{\ell,r_c}^{\top}$ use $r_c(n_\ell+m_\ell)$ entries per module
and include the original LoRA scale $\alpha/r$; do not rescale them when loading.
Truncation can change predictions, so Sec.~\ref{sec:sharing_experiments}
evaluates the accuracy--storage trade-off empirically.

One shared update simplifies distribution and needs no target-specific selection;
merging it into backbone weights removes separate adapter computation.

\paragraph{Backbone as an implicit selector.}
Parameter orthogonality does not by itself imply that foreign components
have little effect on predictions.
We interpret each backbone as responding mainly to the component along
its own Solo update.
For nonzero $\bm\tau_i$, define the orthogonal projection matrix
\begin{equation}
\bm\Pi_i=\frac{\bm\tau_i\bm\tau_i^\top}{\lVert\bm\tau_i\rVert_2^2}.
\label{eq:projection}
\end{equation}
Orthogonal updates satisfy $\bm\Pi_i\bm\tau_S=\bm\tau_i$;
nearly orthogonal updates give an approximate equality when the accumulated
cross terms are small (Supplementary Sec.~\ref{app:projection}).
The projection models the backbone's response for the fixed task and
adapter combinations considered; deployment applies the sum directly.
We compare Exact sum with Solo to measure interference from foreign adapters.
We also compare the leave-one-model-out (LOMO) sum, which omits the target
backbone's adapter, with the unadapted backbone.

\subsection{Source-backbone identification}
\label{sec:identification}
Backbone identification helps model owners investigate suspected unauthorized use.
Given a query adapter $\bm\tau_q$, the task is to identify its training
backbone among candidates.
The training task and tensor layout are known; backbone metadata are missing.

The orientation model in Sec.~\ref{sec:crossgeometry} gives conditions
for cross-backbone cosines to concentrate near zero.
Section~\ref{sec:withingeometry} explains positive mean cosine between
independent runs on one backbone in a local gradient-flow model.
These results motivate comparing the query with same-task reference adapters
and selecting the backbone with the highest average cosine.

For each candidate backbone $i$, retain a nonempty reference set $\mathcal R_i$
of same-task adapters from training runs separate from the query's run.
Using whole-adapter normalization, define
\begin{equation}
\widehat i=\arg\max_i s_i(\bm\tau_q),\quad
s_i(\bm\tau_q)=\frac{1}{|\mathcal R_i|}
\sum_{\bm\tau_v\in\mathcal R_i}
\hat{\bm\tau}_q^\top\hat{\bm\tau}_v
\label{eq:id}
\end{equation}
The mean reference update is not renormalized.
Identification requires no query-time images or backbone execution.
For unknown-backbone rejection, a threshold on $\max_i s_i(\bm\tau_q)$ is
calibrated using separate adapters and held-out backbone folds.

Successful identification still requires the correct-backbone score to
exceed every foreign score, which we test empirically.
Task changes can alter update directions, so we evaluate cross-task
identification separately.

\subsection{Registered-adapter reuse detection}
\label{sec:reuse_method}
Given a combined update and registered adapter weights, the task is to decide
whether each \emph{particular registered adapter} contributed to the combination.
For $M$ nonzero registered adapters, write the combined update as
\begin{equation}
\bm\tau=\sum_{j=1}^{M}b_j\hat{\bm\tau}_j+\bm e.
\label{eq:mixture}
\end{equation}
For a merge $\sum_j a_j\bm\tau_j$,
$b_j=a_j\lVert\bm\tau_j\rVert_2$,
with $b_j=0$ for absent registered adapters.
The dictionary size $M$ need not equal the contributor count.
The residual vector $\bm e$ includes unregistered contributions and
processing errors, which need not be small.
For compressed mixtures, $\bm\tau$ denotes the update after truncated SVD.
The residual $\bm e$ includes its difference from the original sum.

Taking the inner product with registered adapter $i$ gives
\begin{equation}
\widetilde b_i=\hat{\bm\tau}_i^\top\bm\tau
=b_i+\sum_{j\ne i}b_j C_{ij}+\hat{\bm\tau}_i^\top\bm e,
\label{eq:detectionscore}
\end{equation}
where $C_{ij}=\hat{\bm\tau}_i^\top\hat{\bm\tau}_j$.
For same-task adapters trained on distinct backbones,
\cref{thm:orthogonality} bounds these cosines under (A1)--(A2).
Small cross-backbone cosines reduce mixing between registered components.
Detection requires the accumulated cross terms and residual projection to
remain small relative to a present target component.
We use $|\widetilde b_i|/\lVert\bm\tau\rVert_2$
(zero if $\bm\tau=\bm0$) to remove the overall update scale,
and calibrate detection thresholds on held-out mixtures.

Least squares accounts for registered-adapter correlations through the
Gram matrix $\bm C=[C_{ij}]\in\mathbb R^{M\times M}$.
For linearly independent registered updates,
\begin{equation}
\widehat{\bm b}=\bm C^{-1}
\bigl[\hat{\bm\tau}_j^\top\bm\tau\bigr]_{j=1}^M.
\end{equation}
The bracketed vector contains the $M$ inner products.
The score is $|\widehat b_i|$.

The main experiment combines distinct-backbone adapters.
Independently trained same-backbone, same-task adapters may correlate strongly
with a registered adapter and must be tested as target-absent examples.
Identifying a backbone does not establish reuse of a particular adapter.
Section~\ref{sec:reuse} evaluates detection after rank-16 SVD compression.

\paragraph{Efficient, factorization-invariant computation.}
At module $\ell$, write $\Delta\bm{W}_{i,\ell}=c_i\bm{B}_{i,\ell}\bm{A}_{i,\ell}$,
where $c_i=\alpha_i/r_i$. Then
\begin{equation}
\langle\Delta\bm{W}_{i,\ell},\Delta\bm{W}_{j,\ell}\rangle_F
=c_ic_j\operatorname{tr}\!\left[
(\bm{B}_{i,\ell}^\top\bm{B}_{j,\ell})
(\bm{A}_{j,\ell}\bm{A}_{i,\ell}^\top)\right].
\label{eq:factorip}
\end{equation}
Summing across modules gives $\bm\tau_i^\top\bm\tau_j$ without forming dense matrices.
The formula allows different ranks and is invariant to invertible refactorization.
\section{Experiments}
\label{sec:experiments}
\subsection{Shared experimental setup}
\label{sec:shared_setup}

\paragraph{Backbones.}
We use seven ViT-B/16 CLIP backbones~\cite{radford2021clip} distributed
through OpenCLIP~\cite{ilharco2021openclip}: OpenAI, LAION-400M, LAION-2B,
DataComp-1B~\cite{gadre2023datacomp}, DFN-2B~\cite{fang2024dfn}, MetaCLIP-400M,
and MetaCLIP-2.5B~\cite{xu2024metaclip}.
Checkpoint tags, run-to-seed mapping, and training fields are listed in
Appendix~B.

\begin{figure}[t]
\centering
\includegraphics[width=\linewidth]{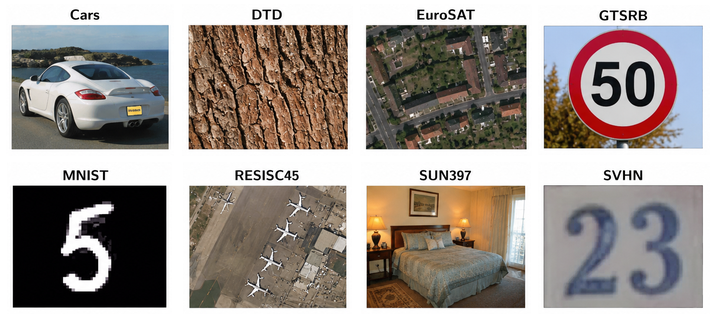}
\caption{Example images from the eight datasets, reproduced from LoRA
superposition~\cite{lora_superposition}.}
\label{fig:datasets}
\end{figure}

\paragraph{Datasets.}
The eight tasks are Cars, DTD, EuroSAT, GTSRB, MNIST, RESISC45, SUN397, and SVHN.
\Cref{fig:datasets} shows example images.
We follow the dataset splits of task arithmetic~\cite{ilharco2023task}
and use common standard OpenCLIP training/evaluation transforms.
Validation images are not used for tuning or early stopping.
Dataset references and split/preprocessing fields are collected in
Appendix~B.

\paragraph{Classification.}
Task-specific classifiers are fixed text prototypes obtained by normalizing
prompt embeddings, averaging across dataset-specific templates,
and normalizing again.
Both the text encoder and the class prototypes remain frozen.

\paragraph{Training.}
Following the Q/V configuration used in the original LoRA study~\cite{hu2022lora},
we apply LoRA only to the query and value projections in all 12 attention blocks,
giving 24 modules with rank $r=16$ and scale $\alpha/r$, $\alpha=32$.
All original backbone parameters remain frozen.
Training minimizes cross-entropy with AdamW, learning rate $10^{-5}$,
weight decay 0.1, batch size 64, and the task-specific epoch budgets in
Appendix~B.
Joint shared matches Solo's epoch budget and optimizer settings.
Each step evaluates the same 64-image mini-batch on all seven backbones,
averages their losses equally, and makes one common update.
The superposition experiments repeat training three times with different
random seeds to assess variability.

\paragraph{Evaluation split.}
The 168 Solo adapters comprise three runs for each backbone--task pair.
Task-matched reference classifiers use one run for queries and the other two
as references.
Detection assigns separate runs to registration, calibration, and testing.
Appendix~B gives the cyclic fold assignments and seed mapping.

\subsection{Update geometry across backbones and tasks}
\Cref{tab:geometry} compares update cosines across distinct runs, backbones, and tasks.
Same-task comparisons test the within- versus cross-backbone similarity margin
for identification.
The same-backbone/same-task condition also tests whether the local model's
positive mean (Sec.~\ref{sec:withingeometry}) persists after AdamW training.
The cross-task condition motivates cross-task identification.
Each condition includes all run pairs (Appendix~B).

\Cref{fig:existing}(a) reports the original mean \emph{absolute} cross-backbone
cosines over eight tasks and three training runs.
The diagonal compares each adapter with itself; similarity across distinct
training runs is assessed separately in \cref{tab:geometry}.

\begin{table}[t]
\centering
\caption{Mean update cosine similarities across backbones and tasks. Self-comparisons are excluded. All entries are computed using the complete run-pair protocol described in Appendix~B.}
\label{tab:geometry}
\setlength{\tabcolsep}{4pt}
\begin{tabular}{llcc}
\toprule
Backbone & Task & Mean cos. & Mean $|$cos.$|$\\
\midrule
Same & Same & 0.424 & 0.424\\
Different & Same & $1.78\times 10^{-4}$& 7.25$\times 10^{-4}$\\
Same & Different & 1.42$\times 10^{-2}$ & 1.45$\times 10^{-2}$\\
\bottomrule
\end{tabular}
\par\vspace{8pt}
\begin{minipage}{\linewidth}
\captionsetup{type=figure}
\centering
\begin{subfigure}[t]{0.485\linewidth}
\includegraphics[width=\linewidth]{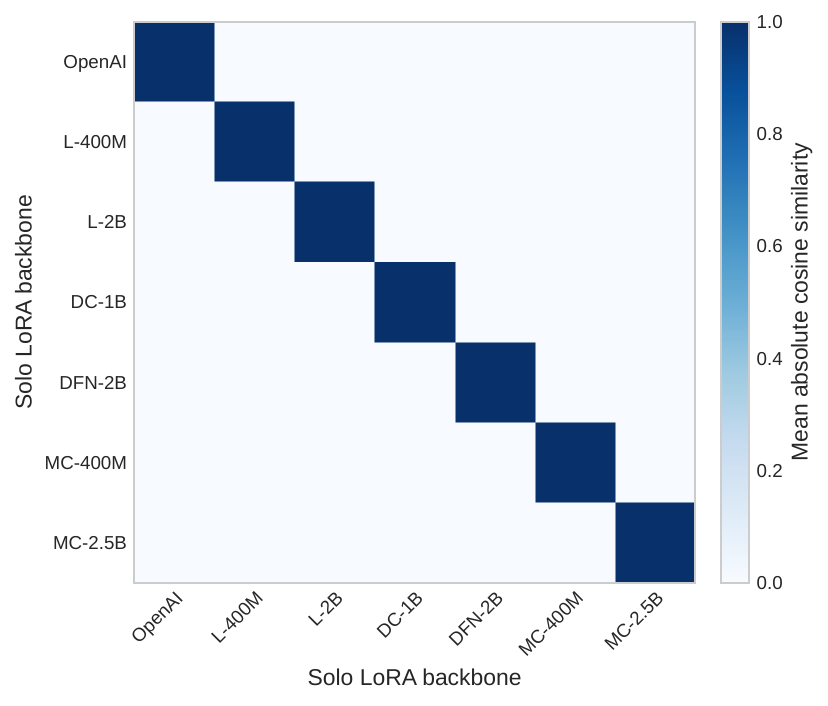}
\caption{Mean absolute cosine.}
\end{subfigure}\hfill
\begin{subfigure}[t]{0.485\linewidth}
\includegraphics[width=\linewidth]{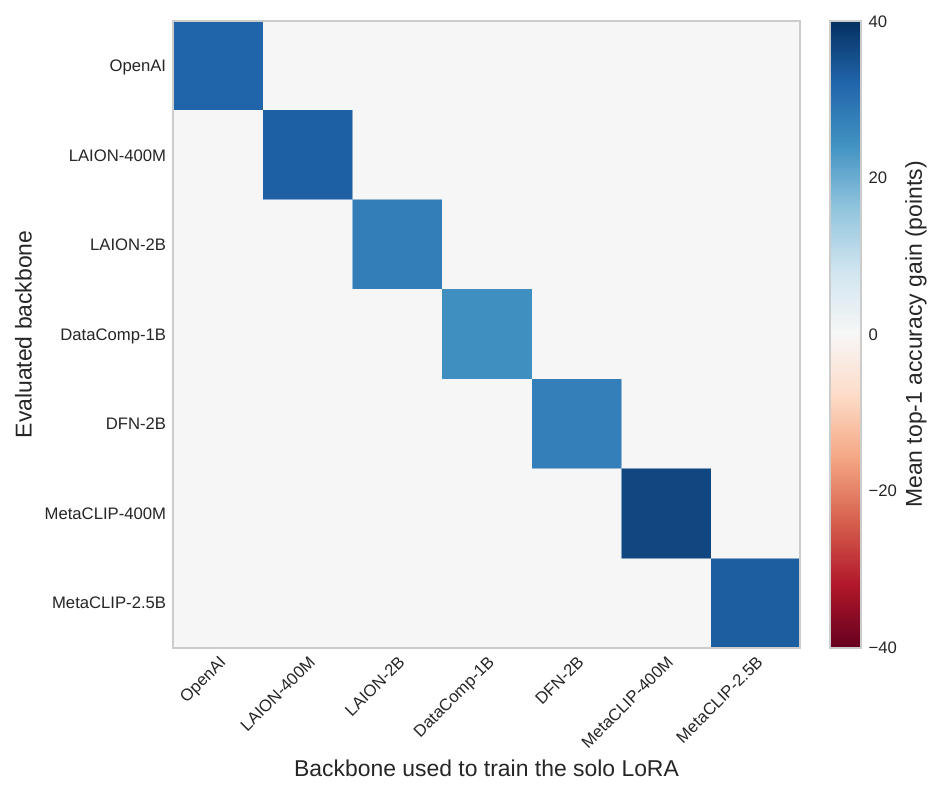}
\caption{Signed accuracy change.}
\end{subfigure}
\caption{LoRA update geometry and cross-backbone response~\cite{lora_superposition}.
Both panels show means over three training runs after averaging eight tasks.
(a) Cosines between LoRAs; the diagonal compares each
adapter with itself. (b) Accuracy gain (percentage points) over the unadapted
row backbone when using a column backbone's LoRA.}
\label{fig:existing}
\end{minipage}
\end{table}

\begin{table*}[t]
\centering
\caption{Top-1 accuracy (\%, three-run mean $\pm$ std)~\cite{lora_superposition}.
\emph{Backbone}: no adapter;
\emph{Solo}: the target backbone's own adapter;
\emph{Exact sum}: sum of seven same-task Solo adapters;
\emph{LOMO sum}: sum of the other six Solo adapters;
\emph{Joint shared}: one adapter jointly trained across all seven backbones.
Per-module ranks: Solo and Joint shared 16; Exact sum at most 112.
Task columns average seven backbones; Average averages eight tasks.}
\label{tab:accuracy}
\setlength{\tabcolsep}{4.1pt}
\begin{tabular}{lccccccccc}
\toprule
Method & Cars & DTD & EuroSAT & GTSRB & MNIST & RESISC & SUN397 & SVHN & Average\\
\midrule
Backbone & \shortstack{$81.82$\\[-1pt]{\scriptsize$\pm0.00$}}&\shortstack{$55.13$\\[-1pt]{\scriptsize$\pm0.00$}}&\shortstack{$54.17$\\[-1pt]{\scriptsize$\pm0.00$}}&\shortstack{$45.12$\\[-1pt]{\scriptsize$\pm0.00$}}&\shortstack{$61.21$\\[-1pt]{\scriptsize$\pm0.00$}}&\shortstack{$67.63$\\[-1pt]{\scriptsize$\pm0.00$}}&\shortstack{$69.10$\\[-1pt]{\scriptsize$\pm0.00$}}&\shortstack{$44.61$\\[-1pt]{\scriptsize$\pm0.00$}}&\shortstack{$59.85$\\[-1pt]{\scriptsize$\pm0.00$}}\\
LOMO sum &\shortstack{$81.72$\\[-1pt]{\scriptsize$\pm0.03$}}&\shortstack{$55.07$\\[-1pt]{\scriptsize$\pm0.09$}}&\shortstack{$54.60$\\[-1pt]{\scriptsize$\pm0.26$}}&\shortstack{$45.27$\\[-1pt]{\scriptsize$\pm0.21$}}&\shortstack{$61.33$\\[-1pt]{\scriptsize$\pm0.33$}}&\shortstack{$67.63$\\[-1pt]{\scriptsize$\pm0.06$}}&\shortstack{$68.53$\\[-1pt]{\scriptsize$\pm0.05$}}&\shortstack{$45.10$\\[-1pt]{\scriptsize$\pm0.13$}}&\shortstack{$59.91$\\[-1pt]{\scriptsize$\pm0.13$}}\\\addlinespace[3pt]
Solo &\shortstack{$85.14$\\[-1pt]{\scriptsize$\pm0.06$}}&\shortstack{$78.54$\\[-1pt]{\scriptsize$\pm0.11$}}&\shortstack{$98.55$\\[-1pt]{\scriptsize$\pm0.04$}}&\shortstack{$96.22$\\[-1pt]{\scriptsize$\pm0.03$}}&\shortstack{$99.09$\\[-1pt]{\scriptsize$\pm0.01$}}&\shortstack{$91.23$\\[-1pt]{\scriptsize$\pm0.04$}}&\shortstack{$79.30$\\[-1pt]{\scriptsize$\pm0.02$}}&\shortstack{$94.67$\\[-1pt]{\scriptsize$\pm0.02$}}&\shortstack{$90.34$\\[-1pt]{\scriptsize$\pm0.02$}}\\
Exact sum &\shortstack{$85.03$\\[-1pt]{\scriptsize$\pm0.13$}}&\shortstack{$77.89$\\[-1pt]{\scriptsize$\pm0.14$}}&\shortstack{$98.49$\\[-1pt]{\scriptsize$\pm0.03$}}&\shortstack{$95.91$\\[-1pt]{\scriptsize$\pm0.03$}}&\shortstack{$99.03$\\[-1pt]{\scriptsize$\pm0.01$}}&\shortstack{$91.03$\\[-1pt]{\scriptsize$\pm0.03$}}&\shortstack{$78.54$\\[-1pt]{\scriptsize$\pm0.02$}}&\shortstack{$94.55$\\[-1pt]{\scriptsize$\pm0.05$}}&\shortstack{$90.06$\\[-1pt]{\scriptsize$\pm0.03$}}\\\addlinespace[3pt]
Joint shared &\shortstack{$84.63$\\[-1pt]{\scriptsize$\pm0.08$}}&\shortstack{$76.68$\\[-1pt]{\scriptsize$\pm0.02$}}&\shortstack{$98.09$\\[-1pt]{\scriptsize$\pm0.03$}}&\shortstack{$94.63$\\[-1pt]{\scriptsize$\pm0.04$}}&\shortstack{$98.74$\\[-1pt]{\scriptsize$\pm0.04$}}&\shortstack{$89.79$\\[-1pt]{\scriptsize$\pm0.07$}}&\shortstack{$78.25$\\[-1pt]{\scriptsize$\pm0.01$}}&\shortstack{$93.96$\\[-1pt]{\scriptsize$\pm0.01$}}&\shortstack{$89.34$\\[-1pt]{\scriptsize$\pm0.02$}}\\
\bottomrule
\end{tabular}
\end{table*}

We also compare observed squared cosines $C_{ij}^2$ with the conditional
second-moment prediction in \cref{sec:theory}.
From the learned updates, we evaluate block-energy weights $w_{i,b}$,
the exact prediction in Supplementary Sec.~\ref{app:moment}, and its bound $\kappa_{ij}$.
An individual squared cosine need not lie below the predicted expectation.
For different-backbone / same-task pairs, the measured mean $C_{ij}^{2}$
matches the prediction within a small factor and stays well below
$\kappa_{ij}$. Random rank-$16$ updates with matching module dimensions and Frobenius
norms provide a baseline for within- and cross-backbone similarities.

\subsection{Application 1: one shared adapter}
\label{sec:sharing_experiments}
\Cref{tab:accuracy} shows little interference in mean accuracy:
Exact sum stays close to Solo despite six foreign adapters.
LOMO sum stays near Backbone, so foreign adapters alone do not recover
the target's adaptation benefit on average.
Together with \cref{fig:existing}(b), this supports the
implicit-selector interpretation (Sec.~\ref{sec:sharing}):
each backbone benefits mainly from its own component in the sum.
Exact sum improves over Backbone in all 56 backbone--task pairs
(three-run means).

The worst Exact--Solo gap is $-1.684$ points (MetaCLIP-2.5B / DTD);
mean absolute foreign-Solo changes are $0.217$ points (Appendix~B).

\paragraph{SVD compression.}
SVD compression (Sec.~\ref{sec:sharing}) retains most of this benefit
(\cref{tab:svd_rank_accuracy}).
Rank 16 reaches 89.55\% accuracy with one Solo adapter's factor storage,
or one seventh of the seven-adapter library.
Higher ranks recover nearly all Exact-sum accuracy at greater storage cost;
more aggressive truncation incurs larger losses.

\begin{table}[t]
\centering
\caption{Test accuracy of Exact sum and its per-module SVD approximations
(Sec.~\ref{sec:sharing}). Entries show means only over 168 equally weighted
backbone--task--run cells. Each task/run sum is compressed separately.
Differences (pp) from Exact sum and Solo use unrounded means.}
\label{tab:svd_rank_accuracy}
\setlength{\tabcolsep}{4pt}
\begin{tabular}{lccc}
\toprule
Method & Accuracy (\%) & $\Delta$ Exact (pp) & $\Delta$ Solo (pp) \\
\midrule
Solo       & 90.34 & $+0.28$  & --- \\
Exact sum  & 90.06 & ---      & $-0.28$ \\
SVD-64     & 90.04 & $-0.01$  & $-0.30$ \\
SVD-32     & 89.95 & $-0.11$  & $-0.39$ \\
SVD-16     & 89.55 & $-0.51$  & $-0.79$ \\
SVD-8      & 88.27 & $-1.79$  & $-2.07$ \\
SVD-4      & 82.66 & $-7.40$  & $-7.68$ \\
SVD-1      & 67.64 & $-22.42$ & $-22.70$ \\
\bottomrule
\end{tabular}
\end{table}

\begin{table*}[t]
\centering
\caption{
Source-backbone identification and unknown-backbone detection.
Task-matched and cross-task results report Top-1 accuracy and
macro-F1 for closed-set identification; Unknown AUROC measures
leave-one-backbone-out detection.
W2T is evaluated using a separately trained weight encoder.
For cross-task identification, it is retrained for each held-out
task and therefore does not use the reference-based protocol.
W2T reports means and population standard deviations over three
encoder-training seeds; AUROC first averages 21 splits within each seed.
}
\label{tab:backbone_identification}
\setlength{\tabcolsep}{6pt}
\begin{tabular}{lccccc}
\toprule
& \multicolumn{2}{c}{Task-matched}
& \multicolumn{1}{c}{Unknown}
& \multicolumn{2}{c}{Cross-task} \\
\cmidrule(lr){2-3}
\cmidrule(lr){4-4}
\cmidrule(lr){5-6}
Method
& Top-1 & Macro-F1
& AUROC
& Top-1 & Macro-F1 \\
\midrule
Log module norms
& 1.000 & 1.000
& $1.000 \pm 0.001$
& 0.262 & 0.257 \\

Singular-value spectra
& 1.000 & 1.000
& $0.986 \pm 0.019$
& 0.238 & 0.241 \\

Module-normalized cosine
& 1.000 & 1.000
& $1.000 \pm 0.000$
& 1.000 & 1.000 \\

Effective-update cosine
& 1.000 & 1.000
& $1.000 \pm 0.000$
& 1.000 & 1.000 \\

W2T
& $1.000 \pm 0.000$ & $1.000 \pm 0.000$
& $0.970 \pm 0.031$
& $1.000 \pm 0.000$ & $1.000 \pm 0.000$ \\
\bottomrule
\end{tabular}
\end{table*}

\subsection{Application 2: source-backbone identification}
We compare \cref{eq:id} with log module Frobenius norms, singular-value
spectra, and mean module cosine.
Each spectrum retains the top 16 singular values and uses per-module $\ell_2$
normalization.
Norm/spectrum features use reference-only z-scores and score each backbone
by its nearest reference in Euclidean distance; both cosine methods average
reference similarities.
A W2T-style encoder~\cite{han2026w2t} learns backbone labels under the same
run split.
Appendix~B details features, permitted training adapters, and encoder training.

\paragraph{Task-matched identification and unknown-backbone detection.}
All methods identify all 168 task-matched queries correctly
(\cref{tab:backbone_identification}).
Backbone signatures therefore persist across independent runs,
but this setting does not distinguish directional scores from norm
or spectral features.

Leaving one backbone out tests rejection when the query's source is absent.
Both cosine scores perfectly separate known and held-out backbones
in the evaluated splits; norms are also nearly perfect.
This supports using maximum cosine to flag absent sources.
W2T's high mean AUROC masks substantial variability:
its minimum is $0.375$ on one of 63 seed--split evaluations.
Its reported statistics summarize three encoder-seed means,
each over 21 splits.

\begin{figure}[t]
\centering
\includegraphics[width=\linewidth]{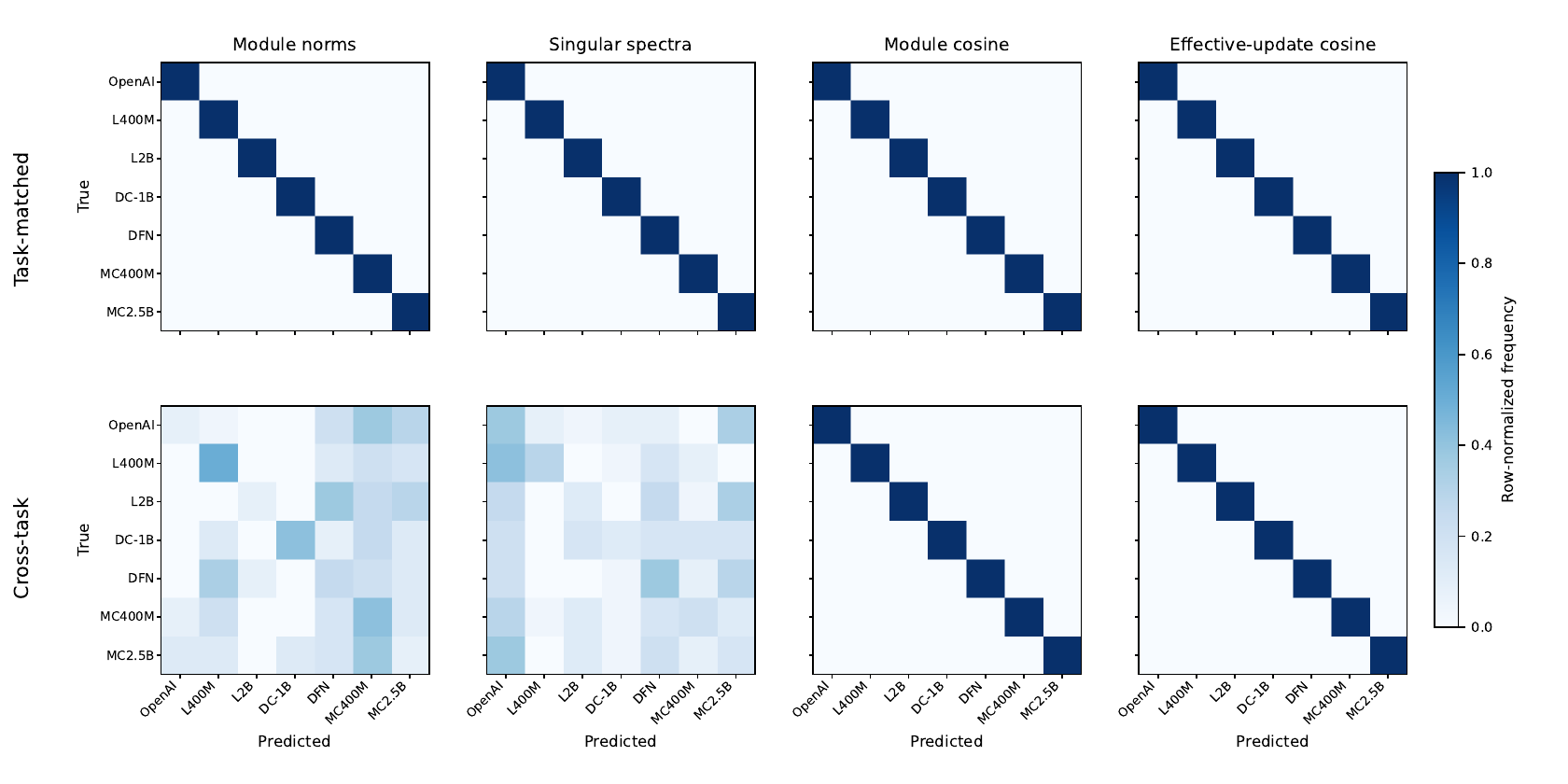}
\caption{Row-normalized confusion matrices for source-backbone identification using four reference-based methods. Top row: task-matched identification. Bottom row: cross-task identification.}
\label{fig:identification_confusion}
\end{figure}

\paragraph{Cross-task identification.}
Removing the query task from references reveals a clearer difference
(\cref{fig:identification_confusion}).
Norm and spectral features lose much of their identification accuracy,
whereas both cosine methods remain perfect.
This supports update directions as source signatures across the evaluated tasks.
W2T also remains perfect after retraining for each held-out task,
with that task excluded from training.
The cosine methods achieve this without a learned encoder.

\subsection{Application 3: detecting a registered adapter}
\label{sec:reuse}
\begin{figure}[t]
\centering\small
\includegraphics[width=\linewidth]{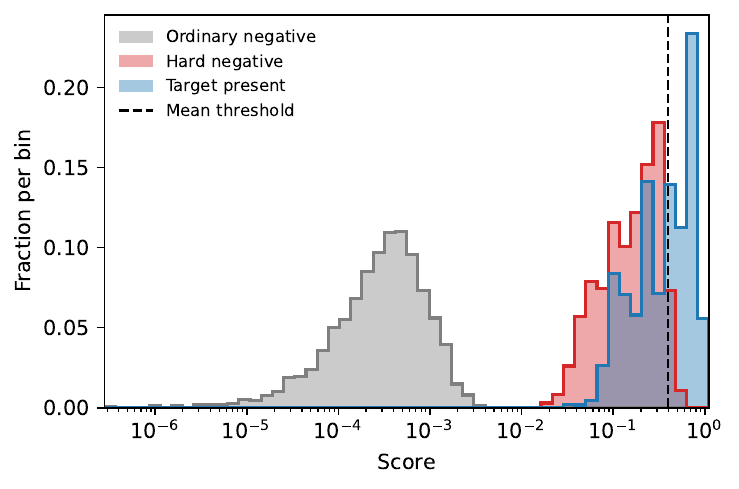}
\caption{Query-normalized detection scores after per-module rank-16 SVD
for target-present mixtures, ordinary negatives, and same-backbone/task
replacements (hard negatives). The dashed line shows the mean of three fold-specific
thresholds, each set to the 95th percentile of SVD-16 hard-negative calibration
scores. Calibration data are separate from test data.}
\label{fig:detection_scores}
\end{figure}

We evaluate mixtures of $K\in\{2,4,7\}$ adapters after per-module
rank-16 truncated SVD (Sec.~\ref{sec:sharing}).
Ordinary negatives remove the target; hard negatives use independent,
norm-matched replacements from the same backbone and task.
\Cref{tab:reuse} shows perfect separation from ordinary negatives.
Query-normalized cosine performs best on hard negatives, but detects only
about half of target-present mixtures at thresholds calibrated on separate
hard negatives to target 5\% FPR.
The scores indicate reuse but do not reliably distinguish registered adapters
from independent same-backbone/task replacements.
\Cref{fig:detection_scores} shows the score distributions after rank-16 SVD;
Appendix~B details the detection protocol.

\begin{table}[t]
\centering\small
\caption{Registered-adapter detection after per-module rank-16 SVD
(3 folds; 6{,}048 positives and 6{,}048 negatives per fold per negative type).
Hard negatives replace the target with an independent same-backbone/task adapter.
Hard TPR uses thresholds set on separate hard-negative calibration data
to target 5\% FPR.}
\label{tab:reuse}
\setlength{\tabcolsep}{3pt}
\begin{tabular}{lccc}
\toprule
Detector & Ordinary & Hard & Hard TPR\\
 & AUROC & AUROC & @cal. 5\% FPR\\
\midrule
Query-normalized cosine & 1.0000 & 0.7790 & 0.520\\
Matched filter          & 1.0000 & 0.7262 & 0.296\\
Joint least squares     & 1.0000 & 0.7262 & 0.296\\
\bottomrule
\end{tabular}
\end{table}

\section{Conclusion}
This study explains how LoRA update directions reflect their training backbones.
A conditional analysis links cross-backbone near-orthogonality to attention
symmetries and equivariant learning.
A local model explains positive mean correlation across independent runs
on the same backbone and task.
Together, these analyses give a theoretical basis for comparing updates
across and within backbones and motivate directional tests of adapter origins.

Experiments show that shared sums retain accuracy close to that of separate
adapters, while low-rank compression reduces storage.
Cosine comparisons identify source backbones across tasks, and inner-product
scores test for registered contributions in combined updates.
These provenance tests use adapter weights directly and require no coordinate
alignment, query-time images, or backbone execution.
A shared update simplifies distribution across models.
Weight-based provenance analysis complements this sharing and helps users
examine adapter origins when metadata are missing.
\clearpage
\section*{LLM Usage Statement}
ChatGPT and Cursor’s coding assistant helped revise the text, check mathematics, and develop figure and analysis code. The authors verified all content and take full responsibility for the paper.
{\small
\bibliographystyle{ieeenat_fullname}
\bibliography{references}
}

\twocolumn[{%
  \pdfbookmark[0]{Supplementary Material}{supplementary}%
  \centering
  {\Large\bfseries Supplementary Material\par}
  \vspace{1em}
}]
\appendix
\setlength{\abovedisplayskip}{7pt plus 2pt minus 3pt}
\setlength{\belowdisplayskip}{7pt plus 2pt minus 3pt}
\setlength{\textfloatsep}{12pt plus 2pt minus 3pt}
\setlength{\dbltextfloatsep}{12pt plus 2pt minus 3pt}
\setcounter{equation}{0}
\setcounter{table}{0}
\renewcommand{\theequation}{S\arabic{equation}}
\renewcommand{\thetable}{S\arabic{table}}
\renewcommand{\theHequation}{supp.\arabic{equation}}
\renewcommand{\theHtable}{supp.\arabic{table}}

Unprefixed equation numbers refer to the main paper.
Supplementary equations and tables use the prefix S.

\section{Proofs and Supplementary Analysis}
\label{app:proof}
\subsection{Attention and LoRA equivariance}
\label{app:equivariance}
For column activations, Eq.~\eqref{eq:gauge} preserves query--key inner products:
$(\bm R\bm q)^\top(\bm R\bm k)=\bm q^\top\bm k$.
Attention probabilities are therefore unchanged, and the output projection
cancels the value transformation:
$(\bm W_O\bm T^\top)(\bm T\bm h)=\bm W_O\bm h$.
The actions are block diagonal across heads, transform Q/K/V biases
consistently, and leave residual coordinates fixed.
For Q LoRA, $\bm B'=\bm R\bm B$ and $\bm A'=\bm A$ give
\begin{equation}
\bm W_Q'+c\bm B'\bm A'=\bm R(\bm W_Q+c\bm B\bm A),
\quad c=\alpha/r.
\end{equation}
The analogous V transformation uses $\bm T$.
The adapted losses agree, so
$\nabla_{\bm B'}\mathcal L'=\bm R\nabla_{\bm B}\mathcal L$
and $\nabla_{\bm A'}\mathcal L'=\nabla_{\bm A}\mathcal L$.
A shared $\bm A$ factor across head blocks is compatible with this action.

For signed permutations, AdamW's first and second moments obey
$\bm m'=\bm R\bm m$ and $\bm v'=|\bm R|\bm v$,
where absolute values are entrywise.
Squaring removes signs, and scalar bias correction preserves both relations.
For bias-corrected moments and a shared stabilizer $\epsilon_{\rm opt}>0$,
\begin{equation}
\frac{\widehat{\bm m}'}{\sqrt{\widehat{\bm v}'}+\epsilon_{\rm opt}}
=\bm R\frac{\widehat{\bm m}}{\sqrt{\widehat{\bm v}}+\epsilon_{\rm opt}},
\end{equation}
with entrywise square roots and division.
Decoupled weight decay also commutes with $\bm R$.
Starting from $\bm B_0=\bm0$, the same $\bm A_0$, and zero optimizer moments,
induction gives $\bm B_k'=\bm R\bm B_k$ and $\bm A_k'=\bm A_k$
at every step; the V case is identical.

Optimizer hyperparameters must be shared across permuted coordinates,
and additional head operations must respect the action.
For the fixed-stream representation below, stochastic operations must
be equivariant for each fixed training stream.
LoRA input dropout on unchanged residual coordinates and
attention-probability dropout satisfy this condition.
Using the same numerical dropout mask on transformed head coordinates
need not, and is excluded.
The argument does not extend to arbitrary rotations under AdamW.

\subsection{From invariant pretraining to an orientation model}
\label{app:orientation}
Let $\mathcal G$ be the product of the finite signed-permutation groups
for all Q/K and V/O head coordinates.
Write $\bm P$ for the induced action on the adapter vector.
Condition on the task and all external pretraining and adaptation choices,
including initialization and stochastic choices for LoRA.
Assume that the conditional joint backbone law is invariant under
an independent $\mathcal G$ action on each backbone.
Sufficient conditions are independent invariant backbone initializations
and pretraining that is equivariant for each fixed stochastic stream,
with the external choices independent of the initial backbone weights.
Pretraining data and objectives may differ across backbones.

Draw a reference collection $(\bm\Theta_i^*)_i$ from this conditional law
and independent uniform $\bm P_i\in\mathcal G$, independently of the references.
Invariance gives
$(\bm P_i\bm\Theta_i^*)_i\overset{d}{=}(\bm\Theta_i)_i$,
where $\overset{d}{=}$ denotes equality in distribution and
$\bm P_i\bm\Theta_i^*$ denotes the action on all backbone parameters.
For fixed adaptation choices $\xi_i$, let
$\bm\tau_i'=\bm\tau(\bm\Theta_i^*,\xi_i)$ be the reference learned update.
Section~\ref{app:equivariance} gives
\begin{equation}
\bm\tau(\bm P_i\bm\Theta_i^*,\xi_i)=\bm P_i\bm\tau_i'.
\end{equation}
Thus Eq.~\eqref{eq:orientation} holds in distribution, with reference updates independent
of the sampled orientations.
The reference updates may be mutually dependent, and $\xi_i$ may be
identical across backbones.
Multiple adapters from one backbone share its one $\bm P_i$.

The construction is a distributional argument, not a coordinate-alignment
step in the applications.
Distinct checkpoint labels do not establish the invariance assumption;
independent orientations can fail for descendants of a common checkpoint.

\subsection{Proof of the moment bound}
\label{app:moment}
Condition on all reference updates.
Reshape $\bm\tau'_{i,b}$ column by column into
$\Delta\bm W'_{i,b}\in\mathbb R^{d_b\times p_b}$,
where $p_b$ is the block's number of columns.
Then $\Delta\bm W_{i,b}=\bm P_{i,b}\Delta\bm W'_{i,b}$.
For one block, put $d=d_b$,
$\bm P=\bm P_{i,b}^\top\bm P_{j,b}$ and
$\bm M=\Delta\bm W'_{j,b}(\Delta\bm W'_{i,b})^\top$.
The relative transformation $\bm P$ is uniform over signed permutations.
Writing $P_{a,\pi(a)}=\sigma_a$, with a uniform permutation $\pi$
and independent equally likely signs $\sigma_a\in\{-1,1\}$, gives
\begin{equation}
Z_b=\bm\tau_{i,b}^\top\bm\tau_{j,b}
=\operatorname{tr}(\bm P\bm M)
=\sum_a\sigma_a M_{\pi(a),a}.
\end{equation}
Averaging first over signs and then over permutations yields
\begin{equation}
\mathbb E[Z_b]=0,
\quad
\mathbb E[Z_b^2]=\frac{\lVert\bm M\rVert_F^2}{d_b}.
\end{equation}
Block orientations are independent, so the second moments add.
The update norms are unchanged by the transformations; hence
\begin{equation}
\mathbb E[C_{ij}^2]=
\frac{\displaystyle\sum_b
\frac{\lVert\Delta\bm W'_{i,b}(\Delta\bm W'_{j,b})^\top\rVert_F^2}{d_b}}
{\lVert\bm\tau_i\rVert_2^2\lVert\bm\tau_j\rVert_2^2}.
\label{eq:exactmoment}
\end{equation}
Using
$\lVert\Delta\bm W'_{i,b}(\Delta\bm W'_{j,b})^\top\rVert_F^2
\leq\lVert\bm\tau_{i,b}\rVert_2^2\lVert\bm\tau_{j,b}\rVert_2^2$
gives Eq.~\eqref{eq:energybound}; Markov's inequality applied to $C_{ij}^2$ gives Eq.~\eqref{eq:tail}.
The product norm in each numerator term is invariant under independent
orthogonal left transformations.
The exact moment and its energy bound can therefore be computed from
observed updates without recovering their orientations.

\subsection{Projection interpretation}
\label{app:projection}
For a fixed task, nonzero Solo updates, and $i\in S$, Eq.~\eqref{eq:projection} gives
\begin{equation}
\begin{split}
\bm\Pi_i\bm\tau_S-\bm\tau_i
&=\bm\Pi_i\bm\tau_{S\setminus\{i\}}\\
&=\left(\sum_{j\in S\setminus\{i\}}
\frac{\lVert\bm\tau_j\rVert_2}{\lVert\bm\tau_i\rVert_2}
C_{ij}\right)\bm\tau_i.
\end{split}
\end{equation}
Exact orthogonality therefore recovers the Solo update from the sum
and gives zero for the LOMO sum.
Near orthogonality gives a small relative projection error when the
norm-weighted sum of cross terms is small.

Let $\bm f_i(\bm x;\bm\tau)$ be the logit vector for input $\bm x$
and update $\bm\tau$.
The idealized selector interpretation additionally assumes
\begin{equation}
\bm f_i(\bm x;\bm\tau)=\bm f_i(\bm x;\bm\Pi_i\bm\tau)
\end{equation}
for the fixed task and adapter combinations considered.
Under this assumption, exact orthogonality gives identical outputs for
Exact sum and Solo, and for LOMO sum and the unadapted backbone.
For nearly orthogonal updates, small output changes also require low
sensitivity to the remaining projection error along the Solo direction.
The response assumption does not follow from parameter orthogonality.
Accuracy comparisons test this interpretation empirically.
Deployment applies the sum without computing a projection.

\subsection{Within-backbone correlation under gradient flow}
\label{app:withinflow}
Fix the backbone, task, training data, and $c=\alpha/r>0$ as in Sec.~\ref{sec:withingeometry}.
Reshape each module's part of the fixed nonzero gradient $\bm g$ into
$\bm G_\ell$; at least one $\bm G_\ell$ is nonzero.
The linearized loss term is
\begin{equation}
\Phi=\bm g^\top\bm\tau
=c\sum_\ell\langle\bm G_\ell,\bm B_\ell\bm A_\ell\rangle_F.
\end{equation}
Its unregularized factor flow is
\begin{equation}
\dot{\bm B}_\ell=-c\bm G_\ell\bm A_\ell^\top,
\quad
\dot{\bm A}_\ell=-c\bm B_\ell^\top\bm G_\ell.
\end{equation}
These linear differential equations have a unique solution at every
finite time.
Each run starts with $\bm B_\ell(0)=\bm0$ and draws the entries of
$\bm A_\ell(0)$ independently from distributions with densities.
Runs are independent and identically distributed.

The chain rule gives
\begin{equation}
\dot\Phi_s(t)=-c^2\sum_\ell\left(
\lVert\bm G_\ell\bm A_{\ell,s}^\top\rVert_F^2+
\lVert\bm B_{\ell,s}^\top\bm G_\ell\rVert_F^2\right)\leq0.
\end{equation}
Each row of $\bm A_\ell(0)$ has a joint density, so it lies in the null space
of a nonzero $\bm G_\ell$ with probability zero.
Hence $\dot\Phi_s(0)<0$ almost surely.
Since $\Phi_s(0)=0$, continuity followed by monotonicity gives
$\Phi_s(t)=\bm g^\top\bm\tau_s(t)<0$ for every finite $t>0$.
The update is therefore nonzero almost surely.

At any fixed $t>0$, the mean normalized update
$\bm\mu(t)=\mathbb E_s[\hat{\bm\tau}_s(t)]$ satisfies
\begin{equation}
-\bm g^\top\bm\mu(t)
=\mathbb E_s\!\left[\frac{-\Phi_s(t)}{\lVert\bm\tau_s(t)\rVert_2}\right]>0.
\end{equation}
The integrand is positive almost surely and bounded by $\lVert\bm g\rVert_2$,
so $\bm\mu(t)\ne\bm0$.
Independent runs at that time give
\begin{equation}
\mathbb E_{s,s'}[\hat{\bm\tau}_s(t)^\top\hat{\bm\tau}_{s'}(t)]
=\lVert\bm\mu(t)\rVert_2^2>0.
\end{equation}
The result guarantees neither a positive cosine for every pair nor a
useful identification margin, and concerns the linearized-loss flow,
not final nonlinear-loss AdamW updates.

The expectation conditions on one backbone and averages over initializations.
A shared orthogonal coordinate change preserves both within-backbone
cosines and $\lVert\bm\mu(t)\rVert_2^2$.
In contrast, Sec.~\ref{sec:crossgeometry} averages independent orientations across backbones.
A shared random orientation can make marginal means zero, but leaves the
two runs dependent; the two results are therefore consistent.

\section{Experimental Details and Controls}
\label{app:protocol}
\subsection{Data, training, and evaluation splits}
\label{app:splits}
\paragraph{Datasets and original sources.}
The datasets are Stanford Cars~\cite{krause2013cars}, DTD~\cite{cimpoi2014dtd},
EuroSAT~\cite{helber2019eurosat}, GTSRB~\cite{stallkamp2011gtsrb},
MNIST~\cite{lecun1998mnist}, RESISC45~\cite{cheng2017resisc45},
SUN397~\cite{xiao2010sun397}, and SVHN~\cite{netzer2011svhn}.
The experiments follow task arithmetic's split convention and do not use
validation images for tuning or early stopping.
All images are processed at $224{\times}224$ with the OpenCLIP transform
of the corresponding pretrained checkpoint (center crop at test time).
Each Solo adapter is trained on the full official or prepared training
split of its task.

\begin{table*}[t]
\centering\small
\caption{Pretrained ViT-B/16 CLIP checkpoints used throughout.}
\label{tab:S1}
\setlength{\tabcolsep}{3pt}
\begin{tabular}{lll}
\toprule
Backbone & OpenCLIP / source tag & Release\\
\midrule
OpenAI
 & \texttt{openai} & OpenCLIP builtin\\
LAION-400M
 & \texttt{laion400m\_e32}
 & OpenCLIP v0.2
 (\texttt{vit\_b\_16-laion400m\_e32-55e67d44.pt})\\
LAION-2B
 & \texttt{laion2b\_s34b\_b88k}
 & LAION CLIP ViT-B/16\\
DataComp-1B
 & \texttt{datacomp\_xl\_s13b\_b90k}
 & DataComp.XL ViT-B/16\\
DFN-2B
 & \texttt{dfn2b}
 & Apple DFN2B CLIP ViT-B/16\\
MetaCLIP-400M
 & \texttt{metaclip\_400m}
 & MetaCLIP B16 400M\\
MetaCLIP-2.5B
 & \texttt{metaclip\_fullcc}
 & MetaCLIP B16 fullcc 2.5B\\
\bottomrule
\end{tabular}
\end{table*}

\paragraph{Checkpoints and training configuration.}
\Cref{tab:S1} lists the seven ViT-B/16 CLIP backbones.
Adapters use LoRA on the query and value projections only
($r{=}16$, $\alpha{=}32$), with $B$ initialized at zero and $A$ by
Kaiming uniform.
Optimization is AdamW with learning rate $10^{-5}$, weight decay $0.1$,
batch size $64$, and a cosine schedule with $500$ warmup steps; label
smoothing is zero and training runs to a fixed epoch budget without early
stopping (Cars $35$, DTD $76$, EuroSAT $12$, GTSRB $11$, MNIST $5$,
RESISC45 $15$, SUN397 $14$, SVHN $4$).
Three independent fine-tunes per backbone--task pair use seeds
$2025$, $42$, and $1234$.
The reported library therefore contains
$7{\times}8{\times}3{=}168$ Solo adapters.

\paragraph{Solo and Joint training budgets.}
The single-process DataLoader supplies $64$ images per step, with no
gradient accumulation; each DataLoader step gives one optimizer update.
Solo evaluates that batch on its own backbone.
Joint shared evaluates the same $64$ images on all seven backbones,
averages the seven losses equally, and updates the common LoRA once.
Solo and Joint use the same task-specific epoch counts, batch size,
AdamW settings, cosine schedule, and $500$-step warmup.
Their computation differs: each Joint update evaluates seven backbones.

\paragraph{Adapter partition.}
Runs 1, 2, and 3 use seeds 2025, 42, and 1234, respectively.
The same role assignment applies across all backbones and tasks.
Identification holds out each run once and uses the other two as references,
giving 56 test adapters per fold and 168 predictions overall.
No threshold is fitted for closed-set identification.

For reuse detection, the ordered triples (registration, calibration,
test) are $(1,2,3)$, $(2,3,1)$, and $(3,1,2)$.
Each run occupies each role once.
Registered targets remain fixed within a fold; calibration and test distractors
and replacements come exclusively from their respective held-out runs.
Modified updates retain their source adapter's role within that fold.
Fold assignments, coefficients, and score choices are fixed before evaluation.
The primary identification and detection experiments require no
additional LoRA training.

\subsection{Geometry and shared-adapter evaluation}
\paragraph{Geometry of Solo updates.}
We measure cosine similarity of effective updates $\tau$ (including the
LoRA scale) over all unordered pairs in three regimes
(\cref{tab:S_geometry}): same backbone and task ($n{=}168$),
different backbones and the same task ($n{=}1512$), and the same
backbone with different tasks ($n{=}1764$).
As a control, we redraw rank-$16$ Gaussian factors for every module,
rescale each module to the learned Frobenius norm, and evaluate the same
pair lists (generator seed $0$).
Same-backbone, same-task updates remain strongly aligned, whereas
different-backbone, same-task pairs lie near the random baseline.
\Cref{tab:S_moment} compares the empirical mean squared cosine with the
block-energy prediction and upper bound from Sec.~\ref{app:moment}.
For different backbones on the same task, the measured second moment
matches the prediction to within a small factor and stays well below the
bound; within-backbone pairs are far more correlated, as expected from
Sec.~\ref{sec:withingeometry}.

\begin{table}[t]
\centering\small
\caption{Mean cosine between Solo effective updates, with a
norm-matched random rank-$16$ control.
Each condition is written as backbone/task
(same or different).}
\label{tab:S_geometry}
\setlength{\tabcolsep}{3pt}
\resizebox{\columnwidth}{!}{
\begin{tabular}{lcccc}
\toprule
Condition (backbone/task)
 & $n$ & mean cos & mean $|$cos$|$ & rand $|$cos$|$\\
\midrule
same/same
 & 168 & $0.424$ & $0.424$ & $2.94{\times}10^{-4}$\\
different/same
 & 1512 & $1.78{\times}10^{-4}$ & $7.25{\times}10^{-4}$ & $2.84{\times}10^{-4}$\\
same/different
 & 1764 & $1.42{\times}10^{-2}$ & $1.45{\times}10^{-2}$ & $2.77{\times}10^{-4}$\\
\bottomrule
\end{tabular}
}
\end{table}

\begin{table}[t]
\centering\small
\caption{Second-moment comparison (pair averages).
Prediction and upper-bound values use the independent-orientation model;
for same-backbone pairs, they serve only as reference values.
Conditions are backbone/task as in \cref{tab:S_geometry}.}
\label{tab:S_moment}
\setlength{\tabcolsep}{2pt}
\resizebox{\columnwidth}{!}{
\begin{tabular}{lccc}
\toprule
Condition (backbone/task)
 & Measured & Prediction & Upper bound\\
\midrule
same/same
 & $1.86{\times}10^{-1}$ & $4.66{\times}10^{-5}$ & $1.26{\times}10^{-4}$\\
different/same
 & $1.28{\times}10^{-6}$ & $1.17{\times}10^{-6}$ & $1.02{\times}10^{-4}$\\
same/different
 & $7.79{\times}10^{-4}$ & $5.77{\times}10^{-6}$ & $9.90{\times}10^{-5}$\\
\bottomrule
\end{tabular}}
\end{table}

\paragraph{SVD aggregation.}
\label{supp:svd_aggregation}
For each task and training run, we sum that run's seven Solo adapters
and truncate each module through thin QR and SVD.
The rank-$k$ approximation is evaluated on all seven backbones using
each task's full test split.
Compression is applied separately to each run's sum, before any averaging
across runs.
We do not fine-tune after composition or compression.
Table~\ref{tab:svd_rank_accuracy} reports an equally weighted mean over all
$7\times8\times3=168$ backbone--task--run cells, with no standard deviation.
Solo and Exact sum use the same full-test aggregation.
Accuracy differences use unrounded means.

\paragraph{Interference and capacity control.}
Across backbone--task cases, Exact minus Solo averages $-0.283$ percentage
points (5th percentile $-0.978$; worst $-1.684$ on MetaCLIP-2.5B / DTD).
A single foreign Solo changes accuracy by $0.217$ points in absolute
value on average.
To match the maximum rank of Exact sum, Joint shared is also trained at
rank $112$ ($\alpha{=}224$, seed $2025$).
Its mean accuracy over the $7\times8$ backbone--task cells is $91.63\%$.
This capacity control uses a single seed, whereas the main Joint
rank-$16$ result averages three seeds.

\subsection{Source-backbone identification}
\paragraph{Reference features and scores.}
\label{supp:id_details}
All reference methods compare the same ordered $24$ $Q$/$V$ modules.
Log module norms use the $24$ features
$\log\lVert\Delta\bm W_m\rVert_F$.
Each norm is the square root of its self-inner-product in the module
Gram matrix; no additive epsilon is used before the logarithm.
Singular-value spectra use the top $16$ singular values of each effective
update $\Delta\bm W_m=(\alpha/r)\bm B_m\bm A_m$, computed through thin QR and SVD.
Each module's spectrum is divided by its $\ell_2$ norm, and the $24$
normalized spectra form a $384$-dimensional feature vector.
For both feature types, each coordinate is z-scored using the mean and
standard deviation of all reference adapters in the fold's candidate set.
A zero standard deviation is replaced by one, and queries use the same
reference statistics.
Each backbone's score is the negative minimum Euclidean distance from
the query to that backbone's references.

Module-normalized cosine first averages the $24$ module cosines for each
query--reference pair and then averages over references of a backbone.
Effective-update cosine averages whole-adapter cosines over references,
as in Eq.~\eqref{eq:id}.
The task-matched setting uses two same-task references per backbone,
one from each non-query run.

\paragraph{Cross-task and unknown-backbone evaluation.}
Cross-task identification removes the query task from the reference bank,
leaving $14$ references per backbone from the two non-query runs.
Unknown-backbone detection excludes each of the seven backbones in turn
from the candidate set, giving $21$ splits across the three run folds.
We do not evaluate near-duplicate checkpoints or other model families
beyond \cref{tab:S1}.

\paragraph{W2T representation and encoder.}
\label{supp:w2t}
We use the original W2T implementation~\cite{han2026w2t} for
QR+SVD canonicalization, the \texttt{FullTransformer} encoder,
AdamW settings, warmup, and sign and rank-permutation augmentations.
The input consists of effective updates $(\alpha/r)\bm B\bm A$.
The original \texttt{canonical\_svd\_features} procedure converts them
to $(\bm u,\bm v,\sigma)$ tokens at $24$ positions
($12$ layers $\times$ query/value), with $16$ rank tokens per position.
The encoder uses hidden width $128$,
\texttt{rank\_encoder\_layers}$=1$,
\texttt{num\_layer\_layers}$=2$, \texttt{nhead}$=4$,
\texttt{mlp\_dim}$=64$, and zero dropout.
The sign and rank-permutation augmentation probabilities are both $0.5$.

\paragraph{W2T training and model selection.}
Closed-set backbone identification uses seven-way cross-entropy,
instead of the multi-label binary cross-entropy used for CelebA/CUB
in the original work.
AdamW uses learning rate $10^{-3}$ and weight decay $10^{-3}$,
with gradient clipping at $1.0$.
Training lasts $100$ epochs with batch size $16$.
The first four epochs update only the projector, embeddings, and classifier;
the remaining $96$ epochs update all parameters with a cosine schedule.
No validation set is used, and the final epoch determines the model.
Relative to the original protocol, we change $45$ epochs to $100$,
batch size $64$ to $16$, validation-best selection to final-epoch
selection, and the original LoRA split to the run folds described below.

\paragraph{W2T evaluation splits and seeds.}
W2T uses the same run split as the reference methods: fold $k$ trains on
runs other than $k$ and tests on run $k$.
For cross-task evaluation, the query task is also excluded from training,
and the encoder is retrained for each held-out task.
For unknown-backbone detection, the excluded backbone is removed from
training, and the unknown score is the negative maximum softmax probability.
Unknown adapters are positive examples.
Encoder seeds $0$, $1$, and $2$ control initialization, data shuffling,
and other W2T training randomness; they are distinct from the LoRA
fine-tuning seeds $2025$, $42$, and $1234$.
Each encoder seed retrains the main, cross-task, and unknown models.

\paragraph{W2T AUROC aggregation.}
Each encoder-seed--fold--excluded-backbone combination gives one AUROC,
with $3\times7=21$ splits per encoder seed.
We first average those $21$ AUROCs within each encoder seed.
The resulting means are approximately $0.926$, $0.988$, and $0.996$.
The reported $0.970\pm0.031$ is their mean and population standard
deviation, as computed by \texttt{numpy.std} with \texttt{ddof=0}.
The minimum $0.375$ is one of the $63$ individual split AUROCs:
encoder seed $0$, fold $3$, and excluded backbone MetaCLIP-2.5B.

\subsection{Registered-adapter detection}
\label{supp:reuse_extra}
The registration, calibration, and test runs follow the partition in
Sec.~\ref{app:splits}.
Each fold uses $M=7$ registered adapters and mixtures of
$K\in\{2,4,7\}$ adapters, with target coefficients
$a\in\{0.25,0.5,1,2\}$.
Ordinary negatives omit the target.
Hard negatives replace it with an independently trained, norm-matched
adapter from the same backbone and task.
For the evaluation in main-paper Table~\ref{tab:reuse} and Figure~\ref{fig:detection_scores}, we first form each mixture
as a weighted sum of adapter updates.
We then apply rank-16 truncated SVD to each module's effective update,
as defined in main-paper Sec.~\ref{sec:sharing}.
All three detectors score this compressed update against the registered
adapters using the definitions in main-paper Sec.~\ref{sec:reuse_method}.
Each fold contains $6{,}048$ positives and $6{,}048$ negatives per negative type.
Hard TPR uses thresholds set on separate hard-negative calibration data
to target $5\%$ FPR.
For query-normalized cosine, the calibration hard-negative mixtures
also undergo rank-16 SVD before scoring.
Each fold's threshold is the 95th percentile of these scores.
Main-paper Figure~\ref{fig:detection_scores} shows the mean of the three calibration thresholds.
Mixtures reuse adapters, so mixture-level observations are not independent.

\end{document}